\documentclass[ijds,fleqn,nonblindrev]{informs4}

\OneAndAHalfSpacedXI

\usepackage{graphicx}%
\usepackage{multirow}%
\usepackage{amsmath, amssymb, amsfonts}%
\usepackage{mathrsfs}%
\usepackage[title]{appendix}%
\usepackage{xcolor}%
\usepackage{textcomp}%
\usepackage{manyfoot}%
\usepackage{booktabs}%
\usepackage{algorithm}%
\usepackage{algorithmicx}%
\usepackage{algpseudocode}%
\usepackage{listings}%
\usepackage{cite}
\usepackage{enumerate}
\usepackage{array}
\usepackage{url} 
\usepackage{hyperref}
\usepackage[hyphenbreaks]{breakurl}

\usepackage[utf8]{inputenc}
\usepackage{csquotes}

\def\tsc#1{\csdef{#1}{\textsc{\lowercase{#1}}\xspace}}
\tsc{WGM}
\tsc{QE}
\tsc{EP}
\tsc{PMS}
\tsc{BEC}
\tsc{DE}

\newcommand{\norm}[1]{\left\lVert#1\right\rVert}

\newcommand{\by}{\times}

\mathchardef\mhyphen="2D

\newcommand {\sS}{\mathcal{S}}

\newcommand {\sX}{\mathcal{X}}

\newcommand \RR {\mathbb{R}}

\usepackage{epstopdf}
\usepackage{subfigure}

\usepackage{natbib}
\bibpunct[, ]{(}{)}{;}{a}{}{,}
\renewcommand\bibfont{\fontsize{10}{12}\selectfont}

\theoremstyle{plain}

\theoremstyle{definition}

\theoremstyle{remark}

\usepackage{natbib}
 \bibpunct[, ]{(}{)}{,}{a}{}{,}%
 \def\bibfont{\small}%
\ECRepeatTheorems

\EquationsNumberedThrough 

\MANUSCRIPTNO{IJDS-0001-1922.65}

\begin{document}



\RUNTITLE{Tis a Butter Place}

\TITLE{Ano-SuPs: Multi-size anomaly detection for manufactured products by identifying suspected patches}

\ARTICLEAUTHORS{%
\AUTHOR{Hao Xu,\textsuperscript{a} Juan Du,\textsuperscript{*,a,b} Andi Wang,\textsuperscript{c} Ying-Cong Chen\textsuperscript{d,e}}
\AFF{\textsuperscript{a}Smart Manufacturing Thrust, Systems Hub, The Hong Kong University of Science and Technology (Guangzhou), Guangzhou, Guangdong, China; 
\textsuperscript{b}Department of Mechanical and Aerospace Engineering, The Hong Kong University of Science and Technology, Hong Kong SAR, Hong Kong SAR, China;
\textsuperscript{c}Department of Industrial and Systems Engineering, College of Engineering, University of Wisconsin-Madison, Madison, USA;
\textsuperscript{d}AI Thrust, Information Hub, The Hong Kong University of Science and Technology (Guangzhou), Guangzhou, Guangdong, China;
\textsuperscript{e}Department of Computer Science and Engineering, The Hong Kong University of Science and Technology, Hong Kong SAR, Hong Kong SAR, China;\\
\textsuperscript{*}Corresponding author, \EMAIL{juandu@hkust-gz.edu.cn}}
}




\ABSTRACT{Image-based systems have gained popularity owing to their capacity to provide rich manufacturing status information, low implementation costs, and high acquisition rates. However, the complexity of the image background and various anomaly patterns pose new challenges to existing matrix decomposition methods, which are inadequate for modeling requirements. Moreover, the uncertainty of the anomaly can cause anomaly contamination problems, making the designed model and method highly susceptible to external disturbances. To address these challenges, we propose a two-step strategy anomaly detection method that detects anomalies by identifying suspected patches (Ano-SuPs). Specifically, we propose to detect the patches with anomalies by reconstructing the input image twice: the first step is to obtain a set of normal patches by removing those suspected patches, and the second step is to use those normal patches to refine the identification of the patches with anomalies. To demonstrate its effectiveness, we evaluate the proposed method systematically through simulation experiments and case studies. We further identified the key parameters and designed steps that impact the model's performance and efficiency. }%

\FUNDING{This work was supported by Guangdong Basic and Applied Basic Research Foundation under Grant No.2023A1515011656, National Natural Science Foundation of China under Grant 72001139 and No. 72371219, Guangzhou-HKUST(GZ) Joint Funding Program under Grant No.2023A03J0651, Guangzhou Municipal Science and Technology Program under Grant No.202201011235.}


 \HISTORY{  }

\maketitle

%

\section{Introduction}

In manufacturing systems, image data frequently provides detailed information about manufacturing status, such as temperature distribution and surface quality \cite{fang2019image}. As image data collection in the production environment becomes more accessible and less expensive \cite{shi2023process}, various manufacturing processes have employed image-based sensing and real-time decision systems for anomaly detection, for quickly inferring the product quality based on the collected image data. With these images, the faults that occur in the manufacturing processes can be timely identified to avoid products with unacceptable quality \cite{du2022tensor}. 

In this paper, we consider the above image-based defect and anomaly identification problem in the context of manufacturing and production systems. Specifically, we consider the common yet challenging scenario where the anomalies may only appear in certain regions of each image. The specific assumptions on the non-anomaly regions and anomalies are as follows: 

\begin{itemize}

\item Even the non-defective regions of the image samples have a certain degree of variations, given the randomness of the disturbance of the data acquisition environment (e.g., lighting and angle) and the stochastic nature of the manufacturing processes. 
\item On the abnormal images, there can be either one or multiple abnormal regions (defects). Besides, these abnormal regions may have different sizes, even within the same image. 

\end{itemize}

\begin{figure}
\centering
\subfigure[Normal image]{%
\resizebox*{4cm}{!}{\includegraphics{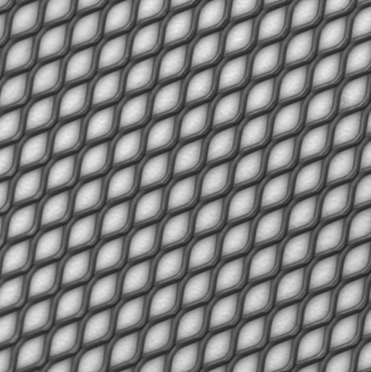}}}\hspace{5pt}
\subfigure[Broken anomaly]{%
\resizebox*{4cm}{!}{\includegraphics{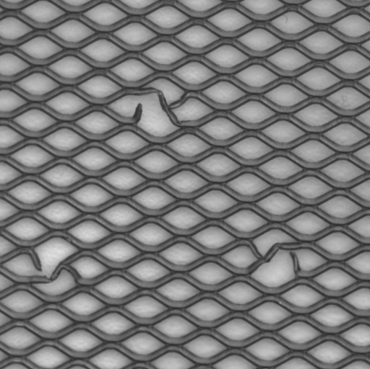}}}\hspace{5pt}
\subfigure[Thread anomaly]{%
\resizebox*{4cm}{!}{\includegraphics{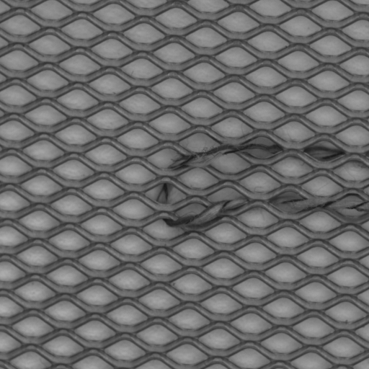}}}
\caption{A metal grid product with different patterns of anomaly from MVTec dataset (a) Normal image with complex backgrounds. (b) Broken anomaly. (c) Thread anomaly.} \label{fig: mesh}
\end{figure}

The following example illustrates the anomaly detection scenario considered in this paper. Three images of metal grid products from the MVTec dataset \cite{bergmann2019MVTec} are displayed in Figure~\ref{fig: mesh}. Among them, Figure~\ref{fig: mesh} (a) shows the image data collected for the normal product, where the surface morphology exhibits a sophisticated geometric pattern, while Figure~\ref{fig: mesh} (b) and (c) shows the image data with anomalies. While these three images have common patterns of metal meshes and the latter two have significant defects, there is a certain level of flexibility in the background of these images, due to the orientation of the metal grid as well as the brightness distribution. As for the anomalies, Figure~\ref{fig: mesh} (b) and (c) display different defective patterns: there are multiple defects in the linkage of the metal grid within Figure~\ref{fig: mesh} (b) while Figure~\ref{fig: mesh} (c) shows a metal grid with a tangled thread caused by process fault. Anomalies from the two latter figures have distinct forms that both lead to non-compliance with the quality standards and the size and the locations of the anomalies are also different.

Local defect identification problems happened in many manufacturing industries, such as semiconductor manufacturing \cite{wang2005using}, textile production\cite{behera2004image}, steel rolling \cite{yan2018real}, and circuit board printing \cite{zhou2023review}. However, anomaly detection based on images obtained from the production systems is a difficult task, given both the variability of the normal regions and the number and sizes of multiple defective regions.

Among existing image monitoring methods, the reconstruction-based method is the most appropriate for our scenario. As will be introduced in the literature review, the existing image anomaly detection methods can be generally divided into two classes of methods: additive matrix decomposition method and deep learning method. Additive matrix decomposition methods typically require stringent requirements that the background is a smooth or low-rank \cite{yan2017anomaly, mou2023paedid} image. In manufacturing practice, however, these assumptions seldom hold. For example, for images in Figure \ref{fig: mesh}, if the orientation of these images is consistent, its image data satisfies the low-rank condition, which can be solved by the additive matrix decomposition method. However, this orientation consistency is difficult to achieve in the real data acquisition process. Thus additive matrix decomposition method needs to be modeled based on specific scenarios and there are limitations on the transferability in real-world applications. For the deep-learning method, image-based anomaly detection can either be based on the latent representation of the image or the pixel-level reconstruction error of the image. The general idea of the reconstruction-based method is to first develop a reconstruction-based model that generates an individual pixel of an image based on neighborhood information. Then the similarity between a pixel and its reconstruction from neighborhood information can be used to identify anomalies. Compared with monitoring the latent representations, the reconstruction-based method is more robust and intuitive.

\begin{figure}
\centering
\subfigure[The first row shows the simulated images from BTAD wood surface data, and the second row shows the corresponding reconstructed images.]{%
\resizebox*{6.85cm}{!}{\includegraphics{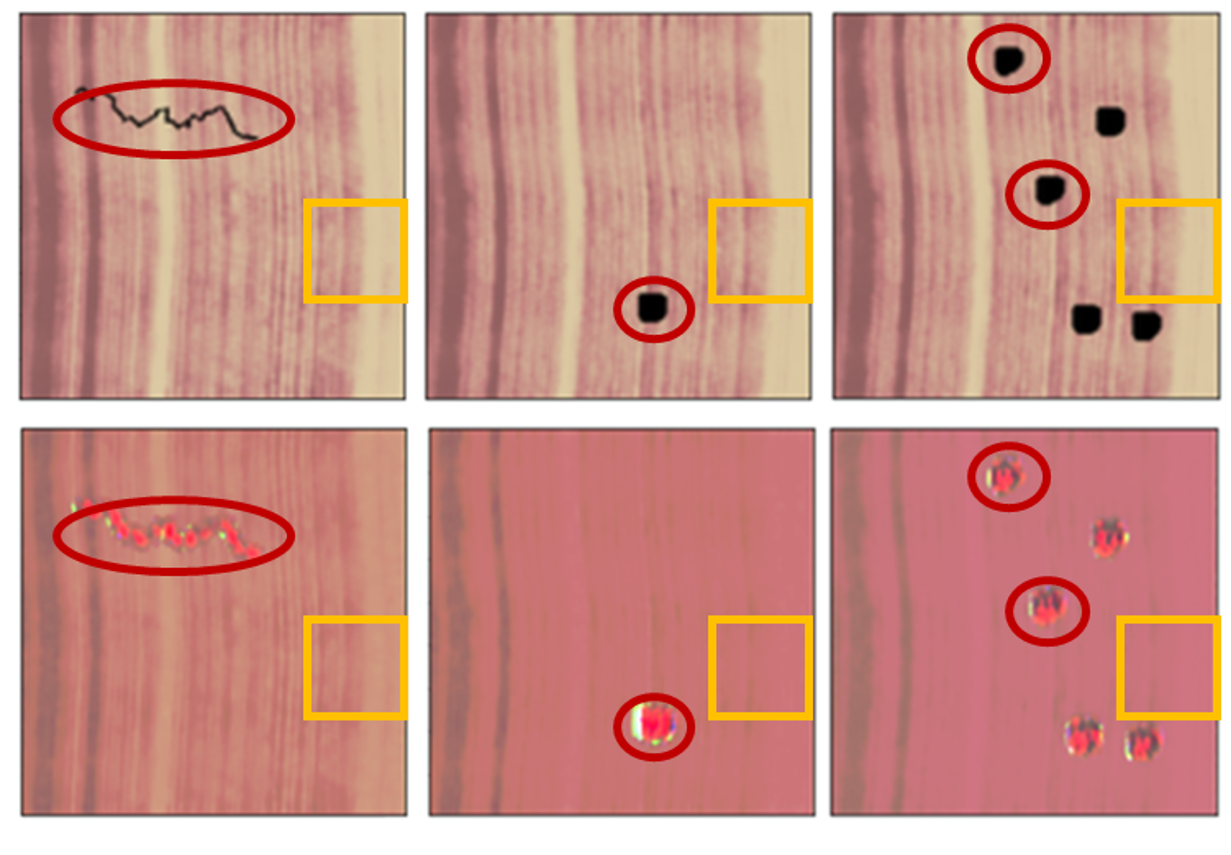}}}\hspace{5pt}
\subfigure[The first row shows the abnormal images from MVTec Hazelnut data, and the second row shows the corresponding reconstructed images. The red area indicates the anomaly part, and the yellow area indicates the normal part.]{%
\resizebox*{7cm}{!}{\includegraphics{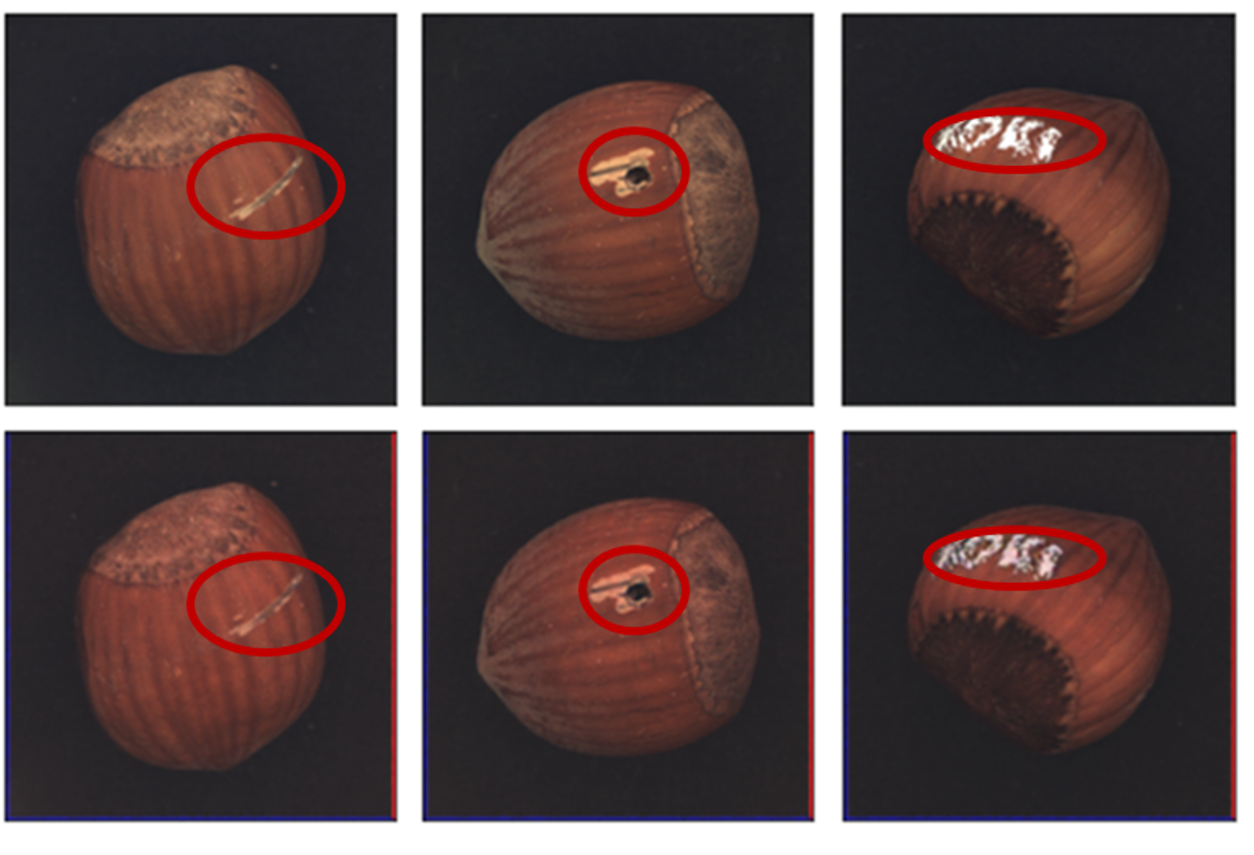}}}
\caption{Two failed reconstruction cases by SSIM-AE}\label{fig: failed reconstruction} 
\end{figure}

However, existing reconstruction-based methods have a common cause that affects the accuracy of anomaly detection. When reconstructing a \textit{normal} patch in an image using other patches that contain anomalies, the reconstruction is influenced by these abnormal inputs, thereby yielding an \textit{abnormal} patch and resulting in imprecise anomaly detection. We refer to this issue as the \textit{anomaly contamination problem}. To illustrate this, Figure~\ref{fig: failed reconstruction} shows two reconstruction cases of SSIM-AE \cite{bergmann2018improving}, a classical reconstruction-based unsupervised anomaly detection method, affected by different anomalies in simulated anomaly images of the MVTec (Hazelnut data from \cite{bergmann2019MVTec}) and BTAD (wood surface data from \cite{mishra2021vt}). In Figure~\ref{fig: failed reconstruction} (a), two different types of anomalies with different sizes were simulated. The comparison revealed that the reconstruction of the anomaly part in the red area and the reconstruction of the normal part in the yellow area based SSIM-AE were affected by the types and sizes of anomalies. Occasionally, due to the powerful learning capabilities of the model, SSIM-AE can even accurately recover the abnormal patches as demonstrated in the red area of Figure~\ref{fig: failed reconstruction} (b), despite that reconstruction intends to infer what the patch should be like \textit{without} anomaly. Such results are especially critical when the image contains anomalies of multiple sizes: since some anomalies may cover multiple patches, certain anomaly patches are surrounded by other patches with anomalies. 

It is also worth mentioning that the anomaly contamination problem cannot be resolved by simply eliminating abnormal samples in the training phase. Even if we can do that, it is still unpredictable what the reconstruction-based model would generate if the input patches contained defects, as no defective patches were provided as training samples in the training phase. No method currently fully addresses the anomaly contamination problem. 

In this paper, we develop an effective and computationally inexpensive image-based anomaly detection method for real-time image monitoring. This method is called \enquote{Anomaly detection through identifying suspected patches} (Ano-SuPs). It is a reconstruction-based method for anomaly detection from images, specifically aimed at resolving the anomaly contamination problem to achieve high anomaly detection accuracy. Specifically, this benefit is achieved by an original two-step strategy. 

\begin{itemize}

 \item In the first step, the possible anomaly patches are identified through a designed patch reconstruction strategy. We reconstruct each patch from several sub-images generated from the input image. Through appropriate strategy, we can adjust the weight of global neighborhood information during the patch-reconstruction process under the Vision Transformer (ViT) structure, enabling the accurate reconstruction of each patch. Based on the reconstruction, we identify the suspected patches where anomalies may occur.
 
 \item The possible anomaly patches identified in the first step may still have errors, given the possibility of having anomaly patches in the reconstruction input. In the second step, we re-evaluate the suspected patches by reconstructing them using the non-suspected patches. Based on the re-evaluation, we identify the actual anomaly patches from the suspected anomaly patches obtained in the first step.
 
\end{itemize}

In the field of industrial image analysis, the proposed method makes the following contributions: 

\begin{enumerate}
\item For reconstruction-based methods in image-based anomaly detection, we design a unique strategy for identifying anomalies based on the advantages of ViT: In the first step of Ano-SuPs, it is demonstrated how the capability of image reconstruction based on random patches can contribute to defect identification of multiple sizes, and thus increases the robustness of the anomaly detection algorithms. 
\item We devised a two-step strategy that can mitigate the anomaly contamination problem to a large extent. After identifying suspected anomaly patches in the first step, no anomaly patches are inputted in the second reconstruction step, enabling contamination-free anomaly detection. 
\end{enumerate}

The rest of this paper is organized as follows. Section \ref{s: works} reviews the related literature on the image-based anomaly detection field. Section \ref{s: methods} presents the proposed method. Section \ref{s: results} provides the corresponding simulation experiments and a case study to evaluate the method's performance. Further analysis of the proposed method to explore the factors that affect the model’s performance is also included. Finally, the conclusions of this paper are in Section \ref{s: conclusion}.

\section{Related work} \label{s: sec2}
\label{s: works}
This section provides a literature review of image-based anomaly detection in industrial applications. We first review the two categories of methods for image-based anomaly detection: additive matrix decomposition methods and deep learning methods. Then, we introduce the ViT, an important analytical tool that we utilize in the Ano-SuPs method. 

\subsection{Additive matrix decomposition methods} \label{s: Matrix decomposition method}

In the field of industrial anomaly detection, the matrix decomposition method has been one of the mainstream methods. The Smooth Sparse Decomposition (SSD) \cite{yan2017anomaly}, Additive Tensor Decomposition (ATD) \cite{mou2021additive} and robust tensor decomposition \cite{shen2022robust} are the prominent examples. These methods assume that the normal region of the image is either smooth or low-rank, while the anomalies on the image are sparse regions. They use optimization problems to characterize the normal regions and the anomalous components. However, in real manufacturing applications, the assumption that the normal region of images is smooth or low-rank is very stringent, which significantly limits the application to the complex pattern and topography of a broader range of manufactured products. Besides, the additive matrix decomposition methods typically involve multiple tuning parameters to specify the trade-off between the requirement of background smoothness and anomaly sparsity, which presents another barrier to the effective and efficient implementation of the anomaly detection algorithm.

\subsection{Deep Learning methods} \label{s: deep-learning methods}

Deep learning has become a dominant modeling method nowadays in many machine learning applications due to its exceptional generalization capabilities and adeptness in handling high-dimensional data \cite{bergmann2018improving, mou2023paedid, raghu2021vision, pirnay2022inpainting, wu2024aekd}. The techniques of using deep learning to perform image-based anomaly detection can be divided into two primary categories \cite{liu2024deep}: representation-based methods and reconstruction-based methods. We review these methods individually and discuss the studies that address the anomaly contamination problem. 

\begin{itemize}
 \item \textbf{Representation-based methods.} The idea of representation-based methods is first to establish the feature extraction function that maps the image samples into latent features and then detects the anomaly based on the discrepancy between the latent features of the current product and the population of latent features of normal products. PaDiM \cite{defard2021padim} applies a pre-trained encoder to extract multi-scale features and construct Gaussian distribution for each pixel. Based on the anomaly scores between the test feature and the learned Gaussian distribution, anomaly detection is performed intuitively and efficiently. However, anomaly detection based on latent representations often suffers from problems such as boundary inaccuracy, which is contrary to the growing need for more precise anomaly information in the manufacturing industry downstream tasks such as root cause analysis. 
 
 \item \textbf{Reconstruction-based methods.} The workflow of reconstruction-based methods is first to establish a prediction function that gives a pixel-level reconstruction of the image based on neighborhood information of the image and then performs anomaly detection based on the reconstruction error. Reconstruction-based anomaly detection methods have the benefit of producing expected patches directly, which are commensurate with the patches of the original image and thereby lead to an intuitive understanding of the defects. 

 Along this line of research, the SSIM-AE \cite{bergmann2018improving} method uses reconstruction error based on the Structural Similarity (SSIM) metric to guide anomaly segmentation, assuming the reconstructed error from the anomaly part will be larger. A Student-teacher network was further proposed in a study by \cite{pirnay2022inpainting}, evaluating both predictive pixel uncertainties and regression error when comparing the output of the student network with that of a pre-trained teacher network trained on large natural images. Some researchers have also tried to enhance the model reconstruction capability by introducing the simulated anomaly samples, such as DRAEM\cite{Zavrtanik_2021_ICCV} and FAIR \cite{liu2023fair}.
\end{itemize}

Only a few works have noticed the anomaly contamination issue in the introduction and attempted to address it formally using the learning algorithms \cite{mou2023paedid, dehaene2020iterative}, yet none of them effectively solve the problem.

\subsection{Vision Transformers} \label{s: vision transformer}

\begin{figure}
\centering
\includegraphics[width=0.9\textwidth]{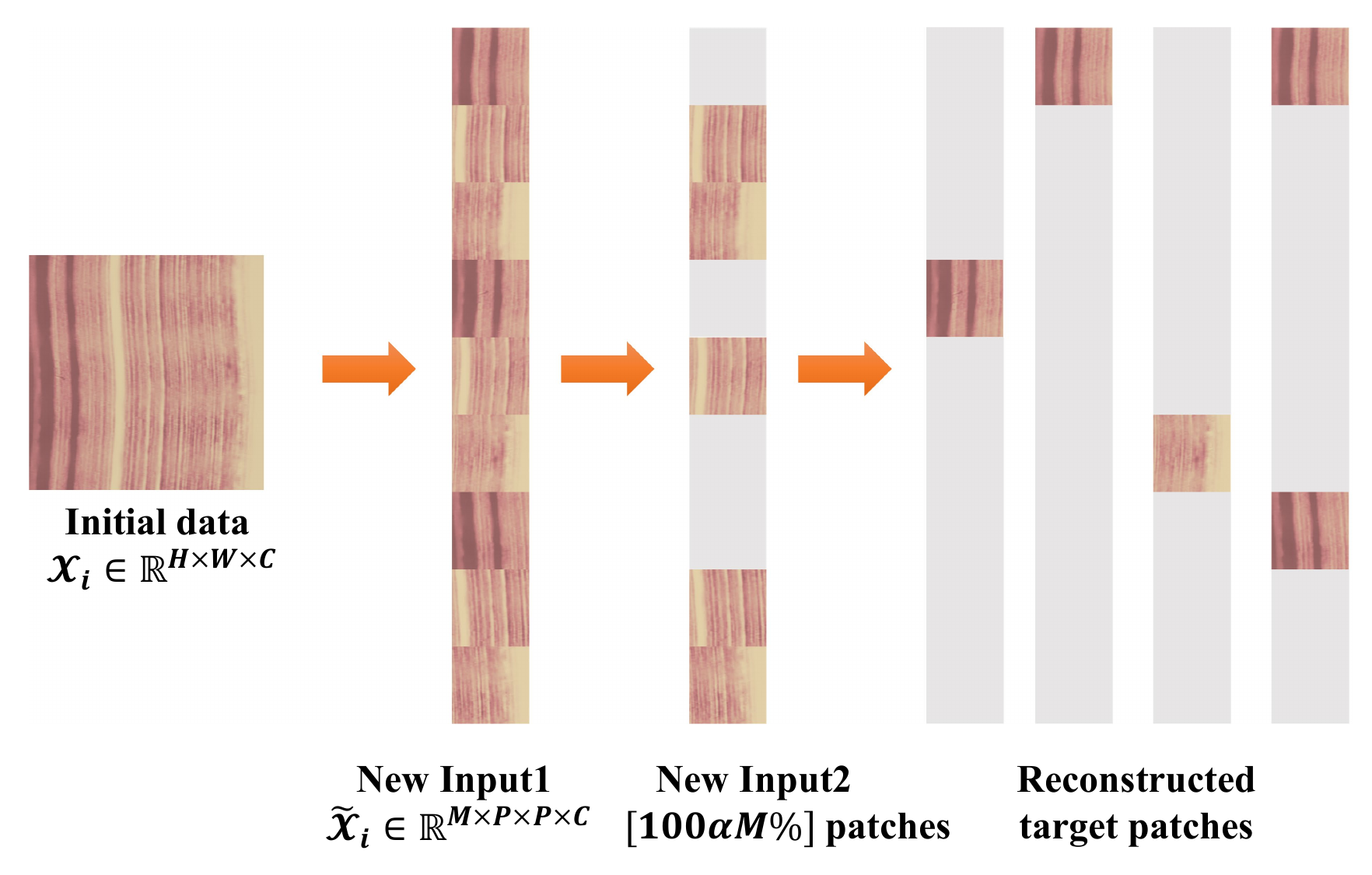}
\caption{An illustration in the information processing of the ViT model using the masked mechanism. The masked patches are filled with zeros.}
\label{fig: ViT}
\end{figure}

In this paper, we develop a patch-reconstruction-based method that eliminates the effect of anomaly. The discussion from the introduction shows the necessity of reconstructing a patch in an image by using patches distributed across the entire image instead of the target patch's neighbors. This goal is explicitly achieved by the ViT model \cite{dosoViTskiy2020image} using the masked mechanism proposed in \cite{he2022masked}. 
ViT is a neural network-based mechanism that transforms images into patches, which help with downstream tasks such as classification and reconstruction. Suppose that we have $N$ training images of size $H\by W$ with $C$ channels, $\{\sX_i \in \RR^{H\by W\by C}: i= 1, \ldots, N$\} and suppose that each patch is a square of size $P\by P$ where $H\by W$ pixels can be divided into $M$ square patches, i.e., $HW=MP^2$. Given the percentage of displaying patches $[\alpha M]$, the training samples of the ViT model are obtained with the following method, as shown in Figure~\ref{fig: ViT}: First, an image $\sX_i$ is randomly selected from the training dataset and transformed into $M$ patches $\tilde{\sX}_i \in \RR^{M\by P\by P\by C}$. From all $M$ patches, $[100\alpha M\%]$ patches are randomly selected as observed patches, and another patch is selected as the target patch of reconstruction. The input data is $\tilde{\sX}_i$ with all unobserved patches set to zeros, and the output is the target patch of reconstruction. The neural network architecture of ViT is designed for the input and output data described above and a transformer mechanism is applied therein to ensure a high accuracy rate of prediction based on relatively small $\alpha$. 

In the literature, few studies used ViT for anomaly detection in images. Among them, \cite{mishra2021vt} used the Gaussian Mixture Model (GMM) to represent the latent vector of ViT, while the detailed anomaly detection method was omitted. \cite{pirnay2022inpainting} still reconstructed patches based on neighbors and calculated discrepancies so the anomaly contamination problem is not resolved. In conclusion, the ViT method offers the possibility of developing a powerful anomaly detection method. Most of the previous work on ViT in anomaly detection has focused on model prediction capability and feature extraction capability. Our approach focuses on the special input-output structure, so our utilization of the ViT for this objective is original.

\section{Proposed method} \label{s: methods}

\begin{figure}[t]
\centering
\includegraphics[width=0.9\textwidth]{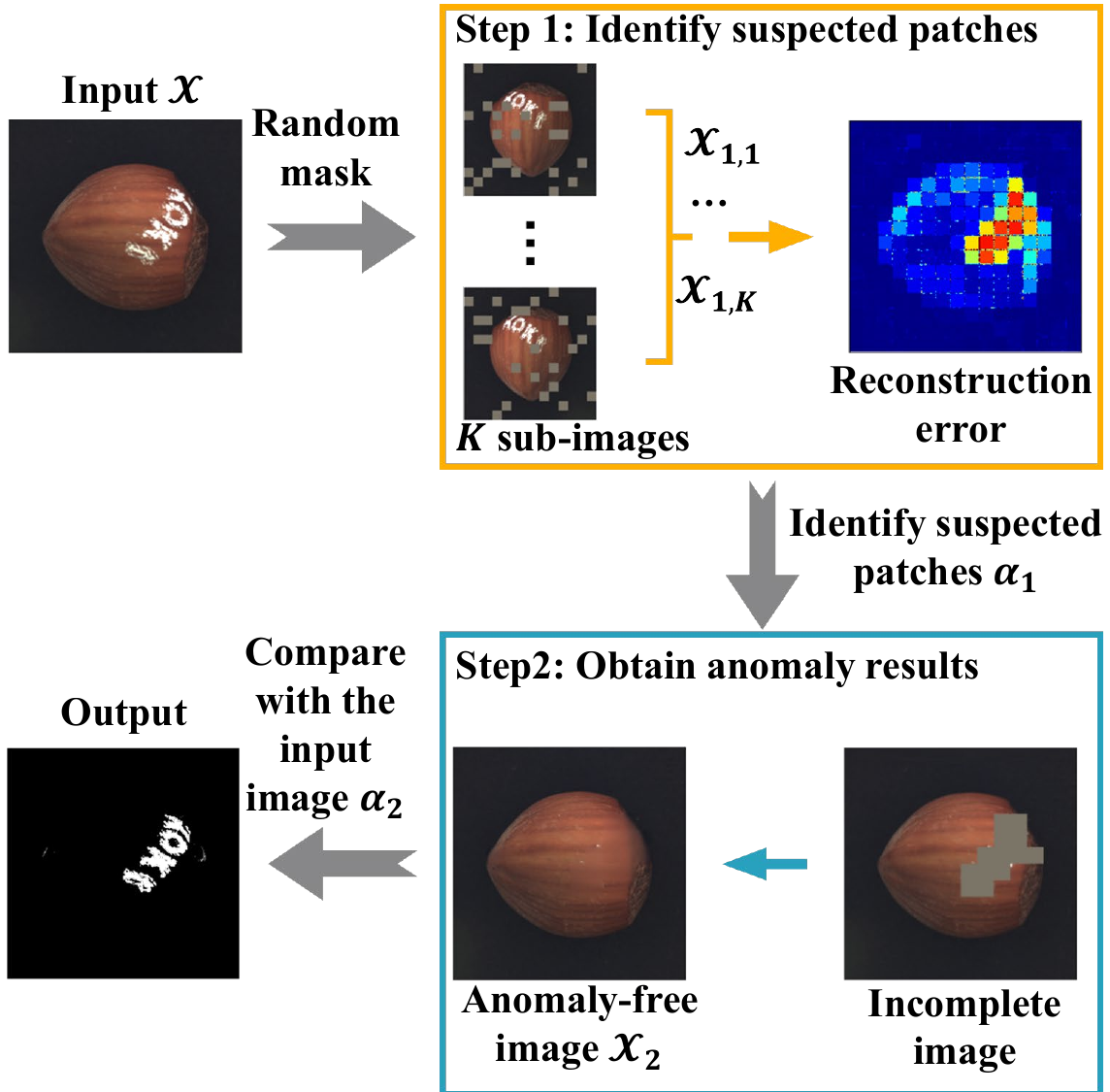}
\caption{Overview of the proposed method}
\label{fig: proposed_method}
\end{figure}

The overview of the proposed method is illustrated in Figure~\ref{fig: proposed_method}. Our Ano-SuPs method is based on finetuning a pre-trained ViT network \cite{he2022masked} to obtain a patch reconstructor. Then we use a two-step procedure for anomaly detection as follows: 
\begin{itemize}
\item[\textbf{Step 1}] We randomly generate $K$ incomplete images based on the test image to reconstruct the randomly masked patch in each incomplete image, and $K$ is a hyperparameter of the method. Then, we reconstruct each patch by utilizing the unmasked patches of that incomplete image. We identify the \textit{suspected patches with anomalies} based on the reconstruction error of this step.
\item[\textbf{Step 2}] We remove all suspected abnormal patches identified from Step 1 from the image being tested and generate an \textit{incomplete image of normal patches}. Based on these normal patches, we reconstruct the suspected patches and compare the construction with the actual suspected patches, to decide whether they are real anomalies.
\end{itemize}

The underlying idea of the two-step procedure is described as follows. In Step 1, we use patches scattered on the entire image to reconstruct the image and thereby the reconstruction of anomaly patches will rely more on global information than on local information close to anomaly patches, resulting in a higher detection rate for anomaly patches. Nevertheless, a normal patch may still be recognized as an anomaly patch in Step 1, due to the possible anomalies in the patches that reconstruct this normal patch. Therefore, the type I error rate (normal patches being recognized as anomalies) is expected to be higher than the type II error rate (anomalies recognized as normal images), which indicates that the patches not labeled as \textit{suspected anomaly patches} are extremely likely to be normal. Since the input of Step 2 only contains the patches with no anomalies, the reconstruction accuracy for normal patches is further enhanced, and thereby both type I and type II errors are improved after Step 2. In essence, Step 1 circumvents the anomaly contamination of Step 2. 

Detailed descriptions of how we train the ViT model and how the two Steps are performed will be introduced in \ref{s: methods.1}, \ref{s: methods.2}, and \ref{s: methods.3}. Finally, we will provide several discussions of the model in \ref{s: methods.4}.

\subsection{\emph{Training patch reconstructor}}
\label{s: methods.1}

To prepare for our Ano-SuPs algorithm, we first train a patch reconstruction model by fine-tuning a pre-trained MAE. This MAE architecture incorporates ViT and thus has the capability of predicting a patch in an image from a fraction of other patches and the pre-trained model is developed by the authors of that paper with ImageNet-1K. However, since ImageNet-1K mainly includes nature images, we fine-tune the reconstructor based on the target dataset. 

The overall fine-tuning method is based on the input and output data format of MAE. Suppose that we have $N$ training images $\mathcal{D}^{\mathrm{train}} = \{\sX_i^{\mathrm{train}}: i=1, \ldots, N\}$. As illustrated in Figure~\ref{fig: ViT}, we generate each training sample of input and output as follows: we randomly pick one image $\sX_i^{\mathrm{train}}$ from the training image data set and transform it into patch representation $\tilde{\sX}_i^{\mathrm{train}}\in \RR^{N\by P\by P \by C}$. We generate the input and output as introduced in the masking scheme of Section~\ref{s: vision transformer} with $\alpha = 1 - 1/K$. These pairs of input and output are fed into the optimizer to update MAE model parameters. 

After the model is trained, we record the reconstruction error of all patches from the validation set. The empirical distribution of these errors will be used to specify the decision boundary of anomaly detection in the following step 1 and 2.

\subsection{\emph{Step 1: Identify suspected patches}} \label{s: methods.2}

\begin{figure}
 \centering
 \includegraphics[width=0.9\textwidth]{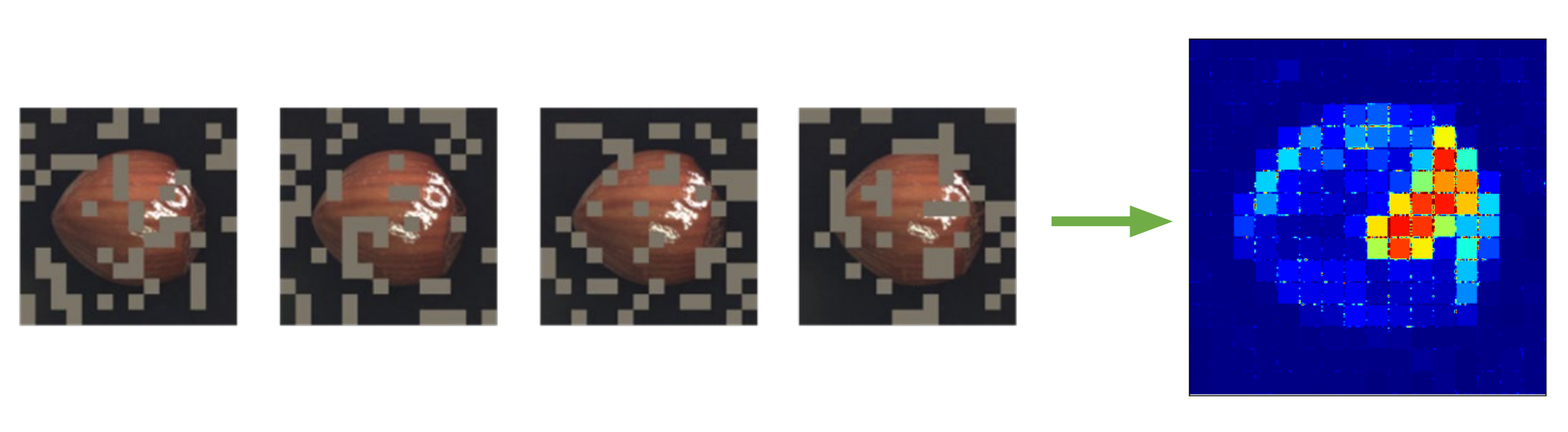}
 \caption{A patch-level reconstruction error visualization example with $K=4$ in Step 1. Yellow color represents patches with small reconstruction errors, red color represents patches with large reconstruction errors.}
 \label{fig: index image}
\end{figure}

From the test image $\sX$, we number all patches $i=1, \ldots, M$ based on their locations and divide them randomly and evenly into $K$ groups of patches. Then we generate $K$ incomplete images ${\sX}_{1, k}, k= 1, \ldots, K$, wherein all patches labeled within group $k$ are removed from the image $k$, $k=1, \ldots, K$. The subscript 1 denotes the incomplete images constructed in Step 1. As a result, all generated incomplete images have approximately $100(1-1/K)\%$ patches being masked, while all the masked patches reconstruct a complete image. An example with $K=4$ incomplete images is shown in Figure~\ref{fig: index image}.

For each patch $i$ of the original image $i=1, \ldots, M$, we identify the generated incomplete image ${\sX}_{1, k(i)}$, where $k(i)$ is the index of the incomplete image within which the patch $i$ is missing. Then we reconstruct the missing patch $i$ within ${\sX}_{1, k(i)}$ using the patch reconstructor obtained from Section~\ref{s: methods.1} and the reconstructed patch is represented as $\hat {\sX}_{1, k(i), i}$. We compare it with the actual patch in the original image $\tilde \sX_{i}$ to calculate the reconstruction error: 
\begin{equation}E_{1, i} = \norm{\hat \sX_{1, k(i), i}-\tilde \sX_i}_F. \end{equation}
Finally, we compare each $E_{1, i}$ with $q_{\alpha_1}$, the upper-$\alpha_1$ sample quantile of the reconstruction error from the testing dataset, and determine that a patch $i$ is a \textit{suspected patch with anomalies} if $E_{1, i} > q_{\alpha_1}$. The output of Step 1 is the set of all suspected patches, given as: 
\begin{equation}\sS = \{i: E_{1, i} > q_{\alpha_1}\}. \end{equation}

\subsection{\emph{Step 2: Obtain anomaly results}} \label{s: methods.3}

After Step 1, we identified the set of suspected patches $\sS$. As discussed at the beginning of this section, the patches with anomalies are only likely to appear in $\sS$. Therefore, we generate another incomplete image $\sX_2$, where all patches in $\sS$ are missing. We then use the same patch reconstructor on individual patch $i\in\sS$ and obtain the reconstructed patch $\hat\sX_{2, i}$. Based on these reconstructed patches, we can calculate the reconstruction errors:
\begin{equation}E_{2, i} = \norm{\hat\sX_{2, i} - \tilde\sX_i}, i\in \sS. \end{equation}
The patches with $E_{2, i} > q_{\alpha_2}$ are decided as containing anomalies.

\subsection{\emph{Discussions}} \label{s: Limitation.5}
\label{s: methods.4}

\textbf{The selection of tuning parameters.\hspace{5pt}} 
There are tuning parameters $K$, $\alpha_1$ and $\alpha_2$. We postpone the discussion on selecting tuning parameters $K$ and the SuPs step in the experiments. The tuning parameters $\alpha_1$ and $\alpha_2$ control the type I error of selecting suspected anomalies and making the final decision on the patches with anomalies. Under the condition that all input patches of the reconstruction model do not have abnormal patches, $\alpha_1$ and $\alpha_2$ approximate the probability that a normal patch is recognized as a suspected patch or an anomaly. Therefore, we suggest choosing both $\alpha_1$ and $\alpha_2$ to be small as long as the patches with anomalies can be identified correctly. 

\textbf{Scope of applicability. \hspace{5pt}} The detailed description of Ano-SuPs indicates the scope of applicability of our method. In short, Ano-SuPs applies to regional anomalies with multiple sizes, where the total area of the anomalies is significantly smaller than the area of the normal regions. In Step 1, the anomaly patches in the inputs are inevitable, as long as the image contains anomalies. The limit of the area of anomaly indicates that the anomaly patches only appear in a small portion of all patches, ensuring that not many patches are influenced by the anomaly contamination problem in Step 1. In this way, abnormal patches can be recognized from their reconstruction (i.e., how the patch \textit{should} look like) and be removed from Step 2. 

\begin{figure}
\centering
\subfigure[Uneven]{%
\resizebox*{6cm}{!}{\includegraphics{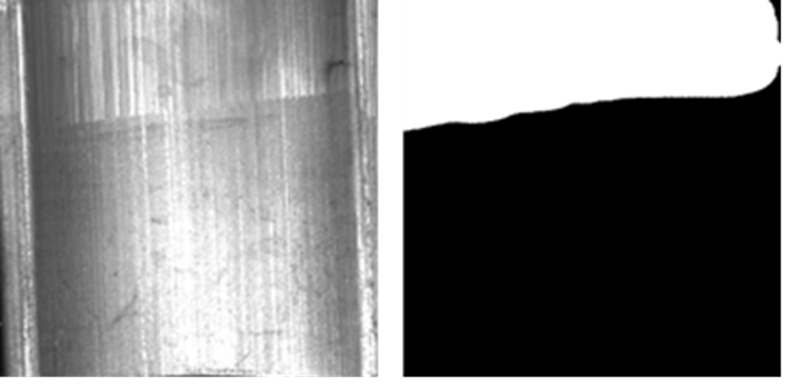}}}\hspace{5pt}
\subfigure[Misplace]{%
\resizebox*{6cm}{!}{\includegraphics{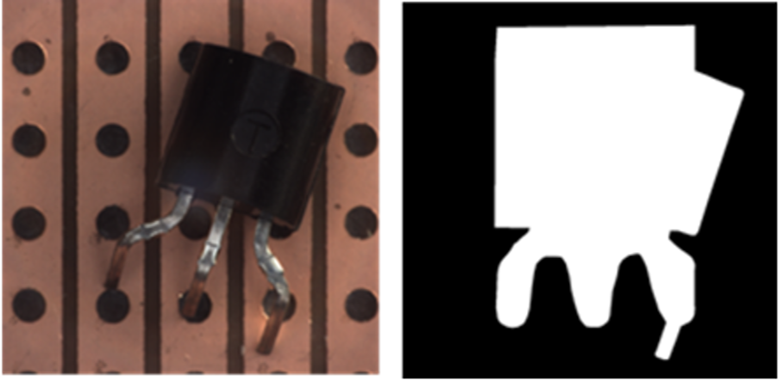}}}
\caption{Some samples that do not satisfy the anomaly assumption based on the proposed method} \label{fig: failed}
\end{figure}

Compared with our scope of applicability, the traditional image-based anomaly detection approaches typically require that the patch size and the defect size be compatible. Ano-SuPs alleviates this constraint and can detect multiple anomalies of different sizes. 

Finally, we illustrate two cases for which the Ano-SuPs approach is not applicable and generates failures. They are illustrated in Figure~\ref{fig: failed}. Both cases contain a large area of anomalies that can be mistaken as being a normal part, which undermines the unsupervised anomaly detection procedure.

\begin{itemize}
 \item Figure~\ref{fig: failed} (a) displays an abnormal sample of a magnetic tile with uneven material distribution in the upper and lower parts of the tile. The anomaly type cannot be detected by our method, because the anomaly is due to the large-scale uneven texture within the image. 
 \item Figure~\ref{fig: failed} (b) displays a sample for an abnormal transistor whose mounting position does not meet the manufacturing standards. Detection of this anomaly requires specific semantic information and thereby is not distinguishable using our method. 
\end{itemize}

\textbf{Insights on Ano-SuPs' advantage. \hspace{5pt}} Besides the scope of applicability, the details of our method also bring insights into how the advantages of our reconstruction-based method are achieved. 

\begin{enumerate} 
 \item \textbf{ViT leads to the effectiveness of screening in Step 1.} When reconstructing the patches in Step 1 based on input patches, the reconstruction of anomaly patches can use the information from other patches potentially far away based on the ViT architecture. For those normal image input patches, the reconstruction error of anomaly will be more noticeable.

 \item \textbf{The screening leads to the effectiveness of anomaly identification of Step 2.} When we reconstruct the anomaly-free image after Step 1, there is no anomaly input in the reconstruction of the second step, which effectively avoids the influence of anomaly in the reconstruction.
 
 \item \textbf{Computational Efficiency.} Once the patch reconstructor is trained, our method only adds one more reconstruction step above the existing patch reconstruction, which does not increase the computational burden. Besides, all $K$ patch reconstruction operations in Step 1 can be processed in parallel. 
\end{enumerate}

\noindent\section{Experiments}\label{s: results}
\begin{figure}
\centering
\subfigure[Two products used in the BTAD dataset.]{%
\resizebox*{5.1cm}{!}{\includegraphics{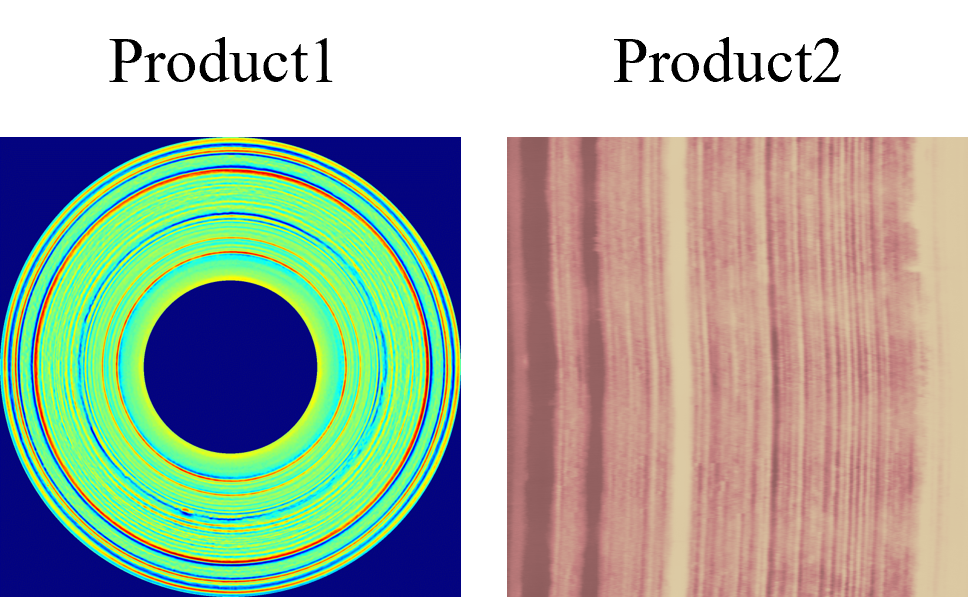}}}\hspace{5pt}
\subfigure[Simulated color anomaly with different sizes and random colors.]{%
\resizebox*{5cm}{!}{\includegraphics{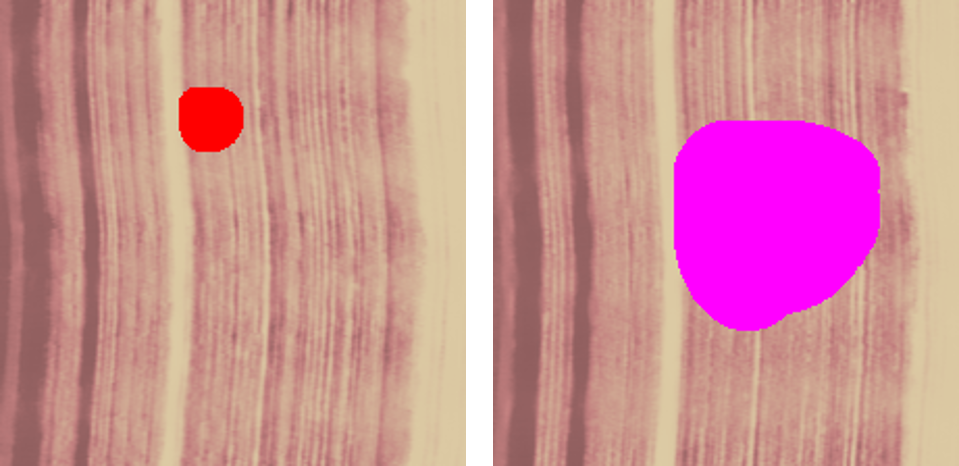}}}\hspace{5pt}
\subfigure[Simulated line anomaly with different lengths.]{%
\resizebox*{5cm}{!}{\includegraphics{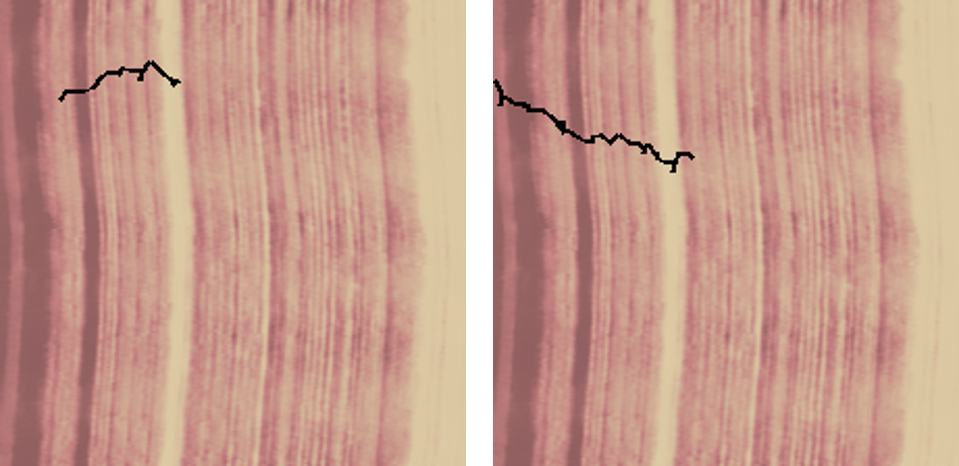}}}\hspace{5pt}
\subfigure[Simulated hole anomaly with varying numbers.]{%
\resizebox*{7.55cm}{!}{\includegraphics{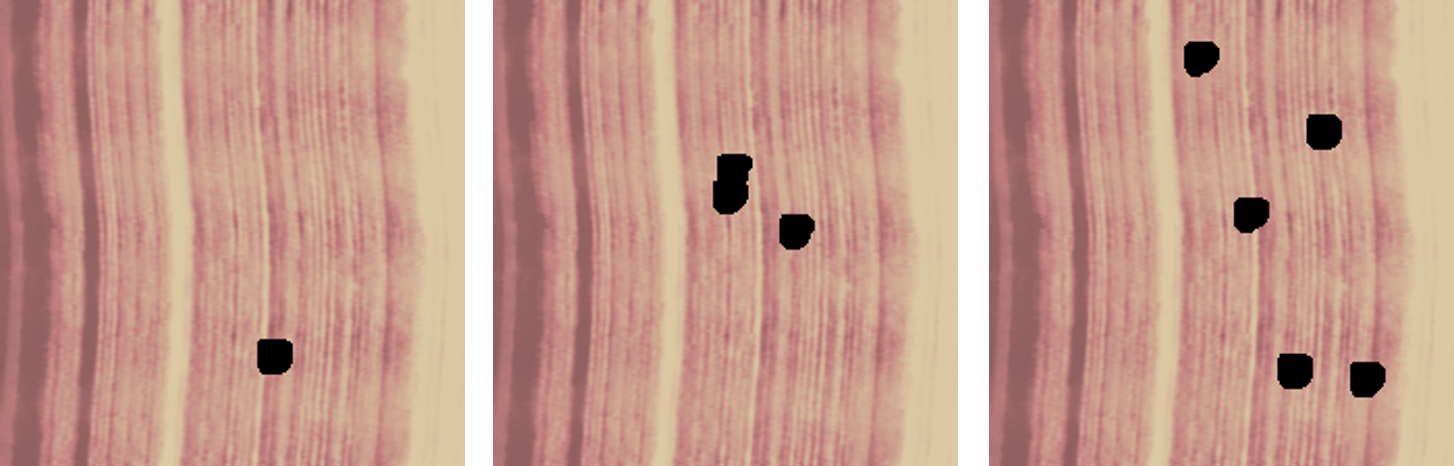}}}
\caption{Some images of products and simulated anomaly images.} \label{fig: Simulation settingl}
\end{figure}

In this section, two experiments are designed to test the performance of Ano-SuPs. Experiments 1 and 2 are based on two BTAD datasets \cite{mishra2021vt} and Experiment 2 is based on MVTec datasets \cite{bergmann2019MVTec}), where the image size of experiments is $224\times224$. In all experiments, we selected $K=2$. Each image is divided into 196 small patches and the size of each patch is $16\times16$. The products without defects in the BTAD dataset and the images generated by the simulation are illustrated in Figure~\ref{fig: Simulation settingl}. We generate and identify three types of anomalies that are common in manufacturing applications: cracks, colored regions, and holes of different sizes.

Experiment 2 is based on the images in the MVTec hazelnut dataset, which comprises four types of anomalies within the images originally: Cut, Crack, Print, and Hole. all these anomalies have different patterns as well as indeterminate sizes.

The study of related parameter $K$ is conducted in an ablation study based on Experiment 2 in Section~\ref{s: ablation}. In all experiments, the decision threshold is selected as the largest reconstruction error of the testing data obtained during the reconstruction model training phase. It corresponds to selecting $\alpha_1$ and $\alpha_2$ as a very small positive number.

\noindent \textbf{Evaluation Criteria.} We use the DICE to evaluate the performance of image-based anomaly detection(\cite{zou2004statistical}). This index is commonly used for assessing the performance of image segmentation and the problem of anomaly detection can be naturally described as segmenting the anomaly region from the normal region of the image. Let $A$ and $B$ be the predicted and the actual defective region, the DICE index is defined as:
\begin{equation} DICE (A, B)=\frac{2 \cdot \mathrm{Area}(A \cap B )}{\mathrm{Area}(A) +\mathrm{Area}(B)}\in [0, 1] \end{equation}
and this index is consistent with the $F1$ score in binary classification problems. A larger DICE indicates better performance of the detection algorithm.

\noindent \textbf{Methods of comparison.} We compare Ano-SuPs with SSIM-AE\cite{bergmann2018improving} and PAEDID\cite{mou2023paedid}. Among them the training of the SSIM-AE is based on the VGG structure and the corresponding threshold value is obtained through cross-validation. The PAEDID method is selected as it also reconstructs an intermediate image to address the anomaly contamination problem. The neural network architecture of PAEDID is described in \cite{mou2023paedid}, and we performed the parameter selection, including the predetermined percentile value (PPV) in the same way as the reference.

\subsection{\emph{Experiment 1: The disk-shaped images and wood surface images in BTAD dataset}} \label{SimuBtad}

\begin{figure}
\centering
\includegraphics[width=0.9\textwidth]{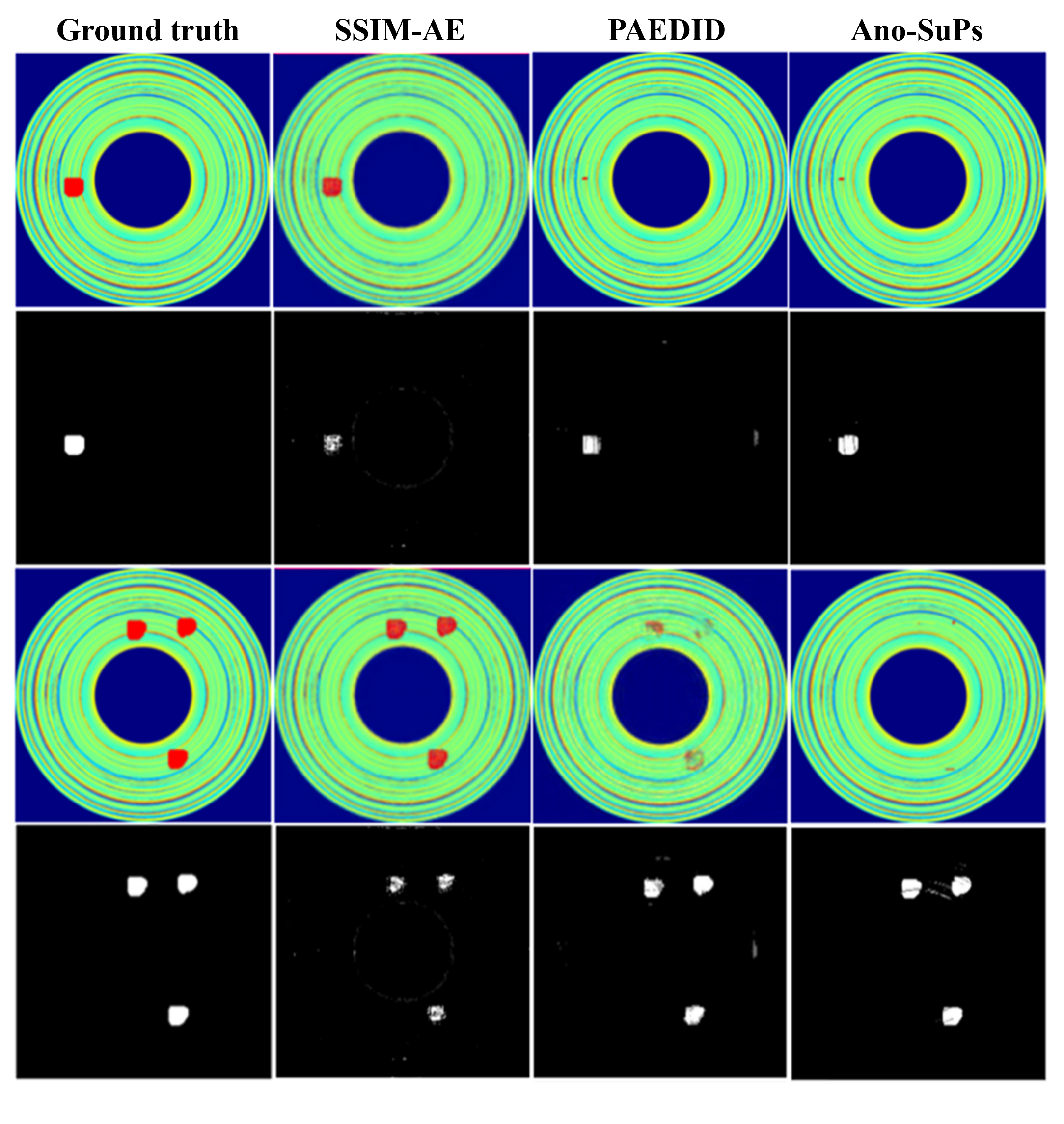}
\caption{ Comparison between the performance of SSIM-AE, PAEDID, and Ano-SuPs for multiple sizes anomaly in disk-shaped images. The first column corresponds to the test image with simulated defective region or Ground Truth images and the rest of the columns correspond to the reconstructed image or anomaly detection results..}
\label{fig: product1}
\end{figure}

\begin{figure}
\centering
\includegraphics[width=0.9\textwidth]{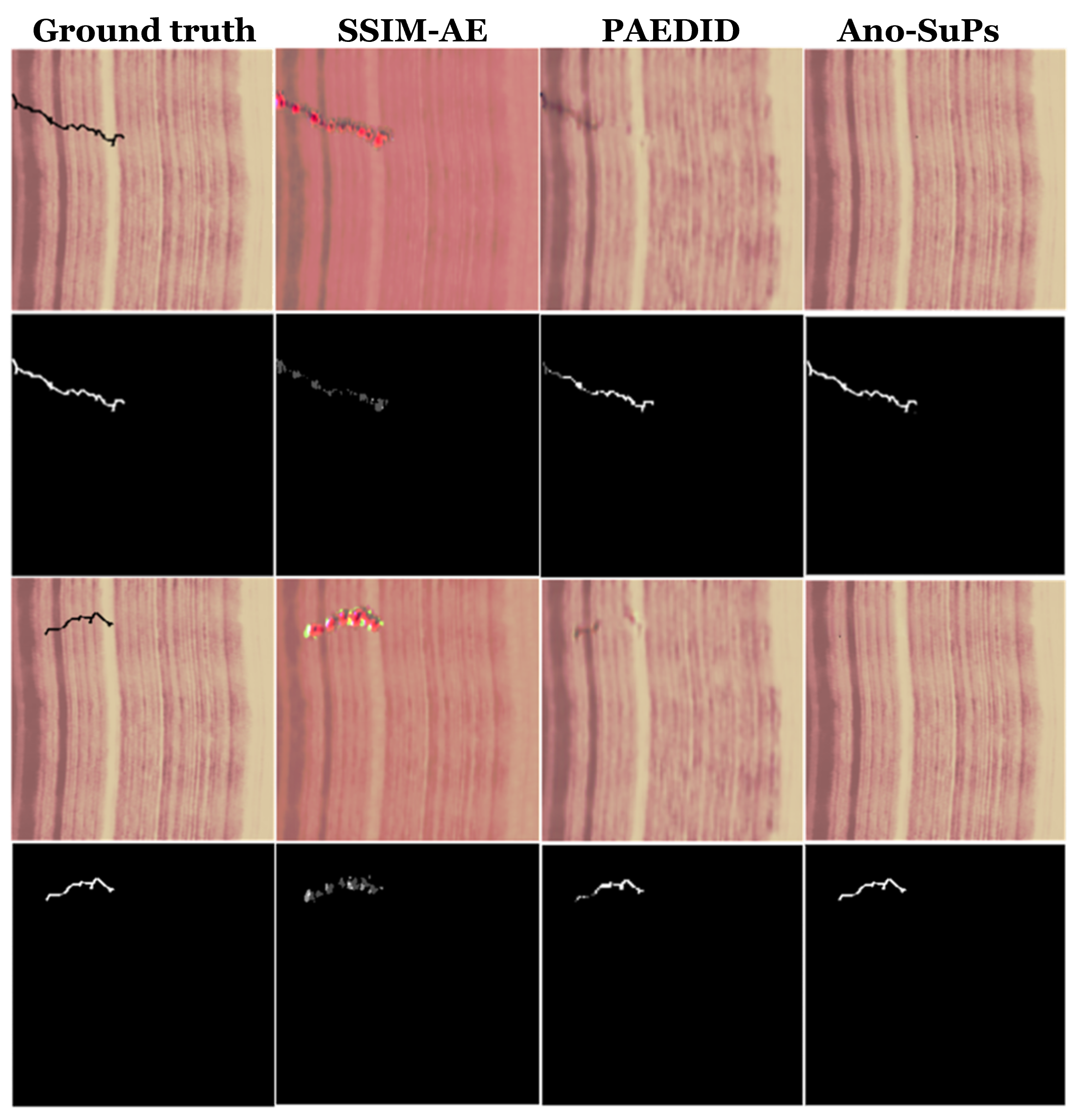}
\caption{ Comparison between the performance of SSIM-AE, PAEDID, and Ano-SuPs for multiple sizes anomaly in wood surface images. The first column corresponds to the test image with simulated defective region or Ground Truth images and the rest of the columns correspond to the reconstructed image or anomaly detection results.}
\label{fig: product2}
\end{figure}

\begin{figure}
\centering
\includegraphics[width=0.9\textwidth]{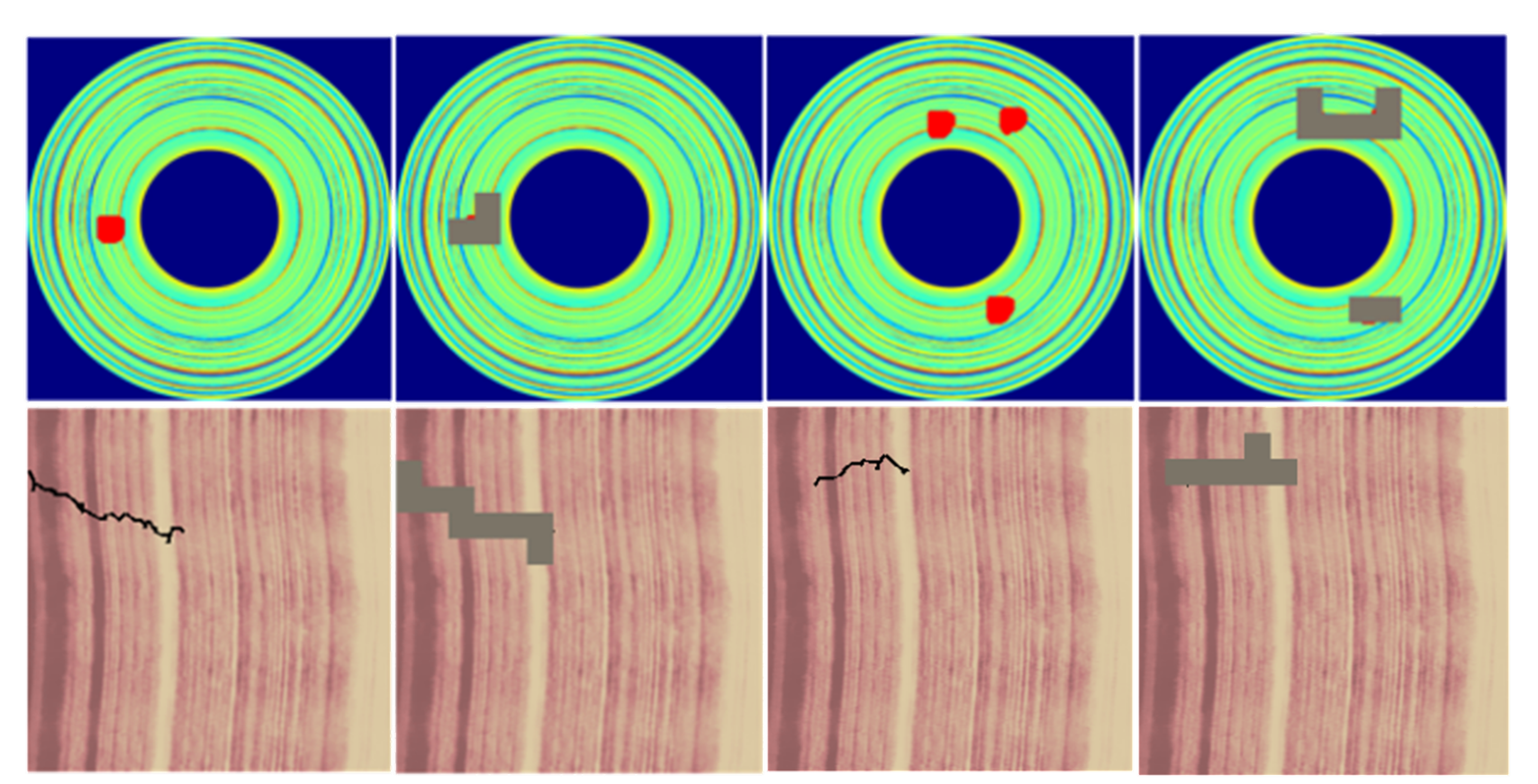}
\caption{ Some examples of incomplete images obtained by Ano-SuPs after Step 1. The first row represents the two test images and the corresponding incomplete image of disk shapes. and the second row represents the two test images and the corresponding incomplete image of wood surfaces.}
\label{fig: incomplete}
\end{figure}

The results of anomaly detection for Experiment 1 are summarized in Table~\ref{tab:SimuBtad}. 
The mean DICE coefficient for multi-sized anomalies in the test images is provided, with the Standard Deviation (STD) of the DICE scores for all generated anomalies enclosed in parentheses. The corresponding incomplete images and detected anomalies using all compared methods are illustrated in Figure~\ref{fig: incomplete}, Figure~\ref{fig: product1} and Figure~\ref{fig: product2}.

For SSIM-AE, the model exhibits two reconstruction failure ways for the disk-shaped images and wood surface images because of anomaly contamination problems. SSIM-AE reconstructs the anomaly regions with certain accuracy (Figure~\ref{fig: product1} and Figure~\ref{fig: product2}) in Product 1 and thus leads to low reconstruction error and poor anomaly detection rate. Meanwhile, the performance of the wood images is poorer than the disk-shaped images because the normal part of the wood images is significantly polluted. We suspect that this is caused by the similarity between the wood texture (especially the stripes on the left side) and the morphology of the crack. Meanwhile, for the stability of the model's performance for multi-size anomalies, its performance differs significantly with the types and sizes of anomalies being detected. The STD of the DICE for hole anomalies is as high as 18.42\% in Product 1 and 15.97\% in Product 2, indicating that the size of the anomaly dramatically impacts the performance of SSIM-AE, due to the anomaly contamination effect.

For PAEDID, the model performance is relatively better than SSIM-AE. The prediction performance is also more stable for different anomaly sizes. The DICE for PAEDID achieves more than 90\% for both color anomalies and hole anomalies. Also, the STD of DICE is lower for PAEDID, indicating that the performance is more stable with various anomaly sizes. The anomaly contamination problem is not significant with a proper PPV, as the reconstructed image contains few anomalies, which confirms the author's conclusions. For example, the reconstructed images in Figure~\ref{fig: product1} and Figure~\ref{fig: product2} demonstrate that PAEDID does not introduce significant errors. However, we can see the trace of a crack within the left-hand side of the wood image and some remain in the disk-shaped image, indicating that the anomaly is not marked fully. This implies room for improvement in the anomaly identification of PAEDID, in terms of overcoming the anomaly contamination problem. 

Ano-SuPs has yielded the best outcomes in Experiment 1, with the mean Dice coefficient exceeding 95\%. It can also be seen from Table~\ref{tab:SimuBtad} that our method has strong stability when dealing with the multi-size anomaly. The STD of the DICE for different sizes of anomalies is also smaller compared to PAEDID and SSIM-AE. The STD of the hole and color anomalies are both smallest among three methods. The illustration of the suspected patches and the reconstructed image in Figure~\ref{fig: incomplete} shows that all patches with anomalies are marked as suspected patches after Step 1 and the reconstructed image after Step 2 does not show any discernable anomalies. 

In general, we have systematically conducted anomaly detection for two different types of products in Experiment 1. After the experiments and analysis, it is proved that Ano-SuPs have stable prediction ability for multi-size anomalies and accomplish the best results.

\begin{table}
\TABLE
{Average DICE and STD for Experiment 1\label{tab:SimuBtad}}
{\begin{tabular}{p{1.4cm}p{1.4cm}p{1.4cm}p{1.4cm}p{1.4cm}p{1.4cm}p{1.5cm}}

\hline
\up & \multicolumn{3}{@{}c@{}}{Product 1} & \multicolumn{3}{@{}c@{}}{Product 2} \down\\
\cmidrule{2-4}\cmidrule{5-7}%
Project & SSIM-AE &PAEDID & Ano-SuPs & SSIM-AE &PAEDID & Ano-SuPs \\
\hline
Line & 6.68\% (2.32\%) &72.66\% (0.17\%) &\bf92.35\% (0.64\%) &28.64\% (4.27\%) &80.74\% (3.38\%) &\bf99.74\% (0.0018\%)\\
Color &63.73\% (3.62\%) &92.14\% (2.16\%) &\bf96.21\% (0.21\%)&86.04\% (6.3\%) &93.44\% (0.53\%) &\bf99.98\% (0.000031\%)\\
Hole &\bf54.79\% (18.42\%) &91.93\% (0.54\%) &\bf94.04\% (0.16\%)&\bf36.73\% (15.97\%) &94.96\% (1.44\%) &\bf99.93\% (0.00016\%)\down\\ \hline
\end{tabular}}
{}
\end{table}

\subsection{\textit{Experiment 2: Anomaly detection in hazelnut images}}\label{s: Case study}

\begin{figure}
\centering
\includegraphics[width=0.9\textwidth]{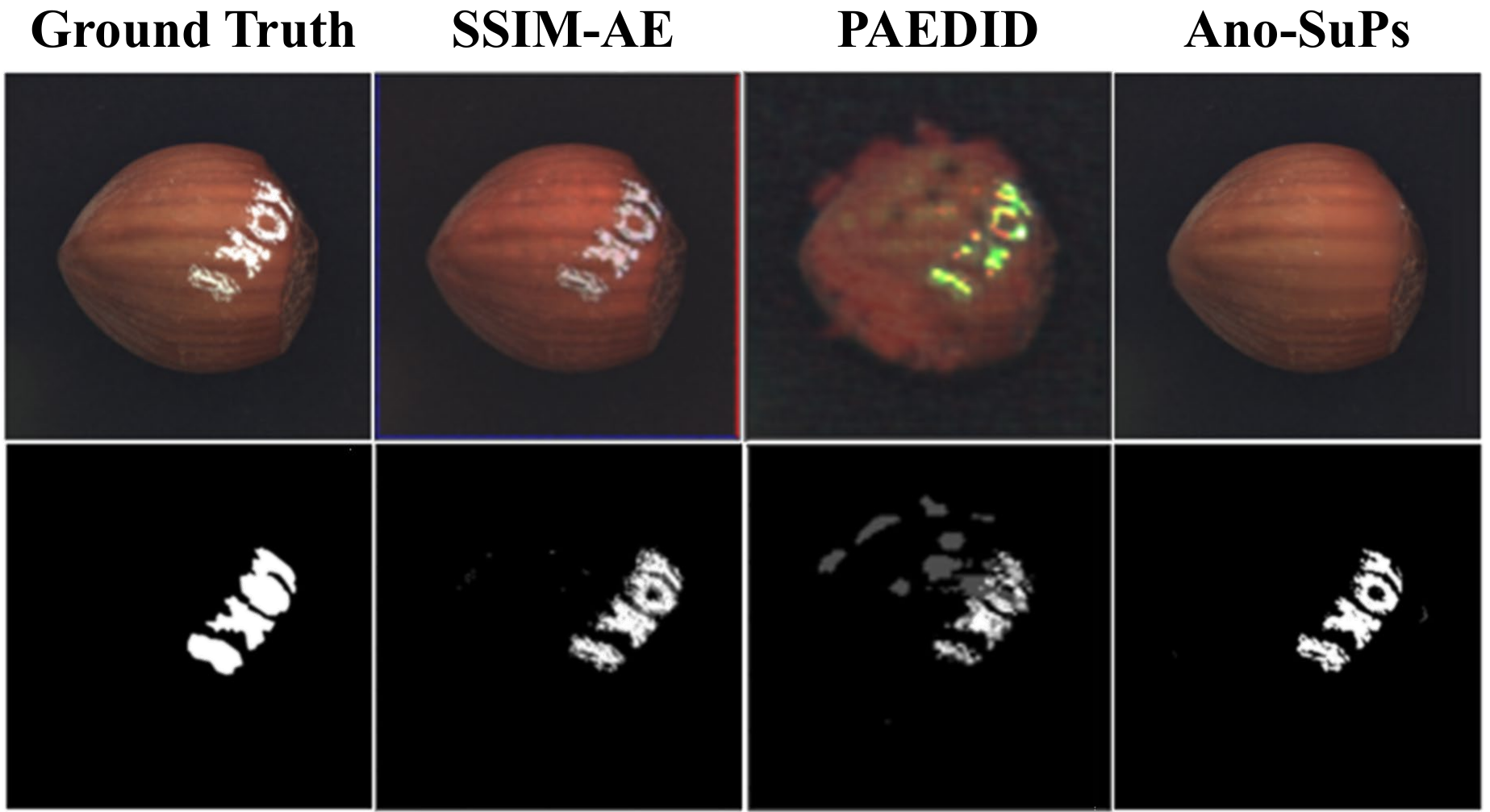}
\caption{Comparison between the performance of SSIM-AE, PAEDID, and Ano-SuPs for the "print" defect in the case study. The first row presents the test image and the reconstructed images for anomaly detection by the different methods, while the second row presents the ground truth defective region and the detected defective region obtained by the different methods.}
\label{fig: case study}
\end{figure}

The corresponding results are shown in Table~\ref{tab:casestudy} and some examples of test images, reconstructed images, and segmentation results of all models regarding print defect are exhibited in Figure~\ref{fig: case study}. For SSIM-AE, due to the anomaly contamination problem, almost all patches with anomalies are well-reconstructed. In contrast, the anomaly part has a relatively large reconstruction error, evidenced by the trace of anomalies in the reconstructed image in the first row.

PAEDID demonstrated a certain performance in this case study. Nonetheless, the reconstructed image reveals that the anomaly component remains present and the normal part has also been impacted, like Experiment 1. This is mainly due to the difficulty in selecting an appropriate PPV value. The PPV value represents the proportion of anomalies replaced in the PAEDID model. In Experiment 2, since the size of anomalies is uncertain, it is difficult to match the PPV value with the real proportion of anomalies for each test sample. A small PPV value will result in incomplete removal of the anomaly, and a large PPV value will result in too many replacements affecting the normal part. Additionally, the hazelnuts in the images have different positions and orientations, which adds to the complexity of anomaly detection. These factors hinder the PAEDID's performance in anomaly detection. 

Our Ano-SuPs method reaches the best results among the three methods for each anomaly type. It successfully identifies the anomaly patches and reconstructs an anomaly-free image, demonstrating its effectiveness in real-world anomaly detection scenarios. Although Step 1 of our method's testing phase is the same as SSIM-AE and the reconstruction error for those anomalous patches is as large as the anomaly region we saw in SSIM-AE, Step 2 effectively avoids the anomaly contamination problem.

\begin{table}
\TABLE
{Average DICE in the Hazelnut case study\label{tab:casestudy}}
{\begin{tabular}{@{}l@{\qquad}l@{\qquad}l@{\qquad}l@{\qquad}}
\hline
\multirow{2}{1cm}{Type} &\multicolumn{3}{c}{Model}\\
\up &PAEDID &SSIM-AE &Ano-SuPs\down \\ 
\hline
\up Cut &14.07\% &32.09\% &\bf44.74\%\\
Crack &32.80\% &28.77\% &\bf47.07\%\\
Print &42.75\% &67.75\% &\bf75.49\%\\
Hole &22.21\% &23.33\% &\bf69.09\%\down\\
\hline
\end{tabular}}
{}
\end{table}
\textit{Ablation studies}\label{s: ablation} To further investigate the impact of designed parameters on the model performance, we conducted an ablation study. Specifically, we focus on the selection of $K$ and the comparison with the method that only uses Step 1 to perform anomaly detection using $E_{1, i}$'s directly. In this ablation study, we compare not only the performance index DICE but also the computational speed, which both factors are likely to affect. To account for the impact of random masks on the model, we conducted 10 repetitions of experiments and calculated both the mean and the STD of the DICE. The results are illustrated in Table~\ref{tab:diss}. 

Our standard method achieves a 59.08\% DICE for all anomalies with per image taking 1.81s.

\begin{table}
\TABLE
{Average DICE and detection time of different factors\label{tab:diss}}
{\begin{tabular}{@{}l@{\qquad}l@{\qquad}l@{\qquad}l@{\qquad}l@{\qquad}l@{\qquad}}
\hline
\up Model &No.images &Time (s) &DICE &STD \down\\
\hline\up 
One-step & 2 &1.16 &45.04\% &0.41\%\\
Ano-SuPs &\bf2 &\bf1.81 &\bf59.08\% &\bf0.47\%\\
Ano-SuPs & 4 &2.99 &57.13\% &0.19\%\\
Ano-SuPs & 8 &5.45 &56.54\% &0.28\%\\
Ano-SuPs & 16 &11.14 &56.34\% &0.33\%\down\\
\hline
\end{tabular}}
{}

\end{table}

In this table, the \enquote{One-step method} refers to the practice of performing anomaly detection directly based on the reconstruction error $E_{1, i}$ after Step 1. To evaluate the impact of the Step 1. The One-step method yielded an average coefficient of 45.04\% and required 1.16 s per image. 
For the STD, the values for all the methods are below 0.5\%, indicating that the random mask mechanism has an impact on the model performance. 
By contrast, when we apply the two-step procedure in Ano-SuPs, there is almost a 12\% improvement in performance, albeit with a corresponding increase in computation times of approximately 0.6s. Our results underscore the importance of the two-step strategy in designing our method.

For the parameter $K$, we systematically studied the results of using $K=4$, 8 and 16. Corresponding results show that the use of more patches yields a gradual decrease in the DICE from 59.08\% to 56.34\% and a significant increase in the testing time cost from 1.81s to 11.14s. 
As the number of $K$ increases, an increasing portion of patches ($1-1/K$) are involved in reconstructing the remaining patches. 
When the number of input patches increases, the reconstruction is increasingly dominated by the neighboring and abnormal patches within the inputs, thereby causing lower reconstruction error for anomaly patches in the first step due to the persisting anomaly contamination problem. 

In summary, we verified that
\begin{itemize}
\item Both step 1 and 2 are indispensable elements of the Ano-SuPs method that yield good performance. The SuPs step before anomaly detection has a huge impact on the model performance, yielding almost a 12\% improvement. 
\item Varying from $K=2$ to 16 leads to a gradual decrease of the DICE from 59.08\% to 56.34\% while a significant increase of testing time from 1.81s to 11.14s. Therefore, we suggest selecting $K=2$ for a better trade-off between performance and computational time. 

\end{itemize}



\section{Conclusion} \label{s: conclusion}

Image-based anomaly detection is a promising research topic in the manufacturing field. However, challenges such as variability and complexity of image background and multi-size anomaly hinder the application in manufacturing scenarios. The anomaly contamination problem we raised shows a research gap between existing methods and the existing image-based anomaly detection research. 

In this article, we propose a two-step Ano-SuPs method to address the above problems and challenges. Our proposed method makes full use of the structural features of ViT, and the designed patch-reconstruction SuPs step does not rely on the neighboring patches, ensuring the determination of multi-size anomalies. The two-step strategy avoids the input of anomalies in the reconstruction of the anomaly-free image, addressing the anomaly contamination problem. Designed simulation experiments, as well as case studies, demonstrate the superiority of our method and confirm the above conclusions. The influence of the parameters and the designed steps is also addressed.

\section*{Related Resources Availability Statement}
The pre-trained model is available at \\ \url{https://dl.fbaipublicfiles.com/mae/pretrain/mae_pretrain_ViT_large.pth}. 

MVTec dataset is available at \\
\url{https://www.mvtec.com/company/research/datasets/mvtec-ad}. 

BTAD dataset is available at\\
\url{https://github.com/pankajmishra000/VT-ADL}.


\bibliographystyle{informs2014} \bibliography{interactcadsample}





\end{document}